\documentclass[10pt,letterpaper]{article}
\usepackage[top=0.85in,left=2.75in,footskip=0.75in]{geometry}

\usepackage{amsmath,amssymb}

\usepackage{changepage}

\usepackage[utf8x]{inputenc}

\usepackage{textcomp,marvosym}

\usepackage{cite}

\usepackage{nameref,hyperref}

\usepackage[right]{lineno}

\usepackage{microtype}
\DisableLigatures[f]{encoding = *, family = * }

\usepackage[table]{xcolor}

\usepackage{array}

\usepackage[aboveskip=1pt,labelfont=bf,labelsep=period,justification=raggedright,singlelinecheck=off, skip=10pt]{caption}

\usepackage{lastpage,fancyhdr,graphicx}
\usepackage{epstopdf}
\usepackage{multirow}

\RequirePackage[singlelinecheck=off]{caption}
\usepackage{xcolor}

\usepackage{placeins}
\newcolumntype{+}{!{\vrule width 2pt}}

\newlength\savedwidth

\raggedright
\makeatletter
\renewcommand{\@biblabel}[1]{\quad#1.}
\makeatother

\fancyheadoffset[L]{2.25in}
\fancyfootoffset[L]{2.25in}
\begin{document}
\vspace*{0.2in}

\begin{flushleft}
    {\Large
        \textbf\newline{A causal learning framework for the analysis and interpretation of COVID-19 clinical data} 
    }
    \newline
    \\
    Elisa Ferrari\textsuperscript{1*},
    Luna Gargani\textsuperscript{2},
    Greta Barbieri\textsuperscript{3},
    Lorenzo Ghiadoni\textsuperscript{3},
    Francesco Faita\textsuperscript{2\Yinyang},
    Davide Bacciu\textsuperscript{4\Yinyang}.
    \\
    \bigskip
    \textbf{1} Scuola Normale Superiore, Pisa, Italy
    \\
    \textbf{2} Institute of Clinical Physiology, C.N.R, Pisa, Italy
    \\
    \textbf{3} Department of Clinical and Experimental Medicine, University of Pisa, Pisa, Italy
    \\
    \textbf{4} Computer Science Department, University of Pisa, Italy
    \\
    \bigskip

    %
    %
    \Yinyang These authors contributed equally to this work.





    * elisa.ferrari@sns.it

\end{flushleft}
\section*{Abstract}

We present a workflow for clinical data analysis that relies on Bayesian Structure Learning (BSL), an unsupervised learning approach, robust to noise and biases, that allows to incorporate prior medical knowledge into the learning process and that provides explainable results in the form of a graph showing the causal connections among the analyzed features. The workflow consists in a multi-step approach that goes from identifying the main causes of patient's outcome through BSL, to the realization of a tool suitable for clinical practice, based on a Binary Decision Tree (BDT), to recognize patients at high-risk with information available already at hospital admission time. 
We evaluate our approach on a feature-rich COVID-19 dataset, showing that the proposed framework provides a schematic overview of the multi-factorial processes that jointly contribute to the outcome. We discuss how these computational findings are confirmed by current understanding of the COVID-19 pathogenesis. Further, our approach yields to a highly interpretable tool correctly predicting the outcome of 85\% of subjects based exclusively on 3 features: age, a previous history of chronic obstructive pulmonary disease and the PaO2/FiO2 ratio at the time of arrival to the hospital. The inclusion of additional information from 4 routine blood tests (Creatinine, Glucose, pO2 and Sodium) increases predictive accuracy to 94.5\%.


\section{Introduction}
Coronavirus disease (COVID-19) first appeared in China in November 2019 and it is caused by the new coronavirus SARS-CoV-2. Its spreading was immediately very fast, prompting WHO to declare a pandemic status on March 12, 2020.
Currently, COVID-19 counts over 116M accumulated cases worldwide with more than 2.5M deaths~\cite{whoCovid}.
Early identification of critical patients is an imperative challenge for triage systems as the severity of cases is putting great pressure on hospital resources, leading to the need of alleviating the shortage of medical assets ~\cite{kirkpatrick2020scarce}. This is especially important during COVID-19 pandemic when the access to the relatively less available supportive care and intensive care units (ICUs) is more frequent and timely interventions may reduce mortality.

Since the pandemic outbreak, several algorithms have been proposed to evaluate predictive scores helping in management of COVID-19 patients. Not surprisingly, most of the proposed tools relied on complex sets of clinical parameters or laboratory markers with only few of them shared between different approaches. These systems are based on both standard statistical approaches and Machine Learning (ML) methodologies~\cite{abd2020artificial}.  The application of supervised ML algorithms to medical research has become widespread in the last years. It has been therefore natural to see a continuation of this trend in the analysis of COVID-19 clinical data, also thanks to the generally superior predictive performance of ML-based solutions over classical statistical approaches. However, the adoption of ML models is also known to be subjected to many pitfalls due to their black box and data-driven nature. In fact, the feckless application of supervised ML often results in overfitting problems, confounding bias effects and scarce interpretability; the latter being particularly troublesome for clinical practice. These phenomena have effectively been noted by a recent meta-analysis~\cite{wynants2020prediction} where predictive models for COVID-19 are defined to be “poorly reported, at high risk of bias, and their reported performance is probably optimistic”.

Another critical issue that needs careful consideration is the fact that commonly adopted ML approaches to diagnosis or outcome prediction are almost uniquely from the family of the supervised learning models. As such, they work on a purely associative approach, by identifying correlations between input data, such as symptoms, clinical history and treatments, and the target variable. Many examples show that the inability to distinguish correlation from causation can produce classifications that are misleading to the point of being potentially dangerous \cite{richens2020improving}. For instance, the paper by Caruana {\it et al.} \cite{caruana2015intelligible} describes a model to predict the probability of death by pneumonia, trained on medical records of patients who have previously had pneumonia. Counter-intuitively, the model found that asthma lowers the risk of death, while it is known to be a severe condition in subjects with pneumonia. This misleading effect occurred because the patients with asthma in the training set received more care by the hospital system. In this example, the association learned from the dataset was correct, but clearly the aim of this application was to find only causal relations useful to prioritize care for patients with pneumonia. 

A good option to avoid this kind of problems is given by Bayesian Structure Learning (BSL) approaches\cite{Spirtes2000}, which allow to learn from data a probabilistic directed acyclic graph (DAG), called Bayesian Network (BN), connecting the elements that according to the Bayes' rule have a causal relationship. With respect to supervised ML models, BSL has several features making it suitable for medical research \cite{Aliferis2003371,miBN}, such as:
\begin{itemize}
    \item it is intrinsically interpretable given that it produces a graph representing the causal relationships among the input data;
    \item it allows to include domain knowledge \cite{uusitalo2007advantages}, i.e. it can take into account a set of a-priori known mandatory and prohibited connections. This is a useful feature to combine consolidated medical knowledge with the ongoing research;
    \item it naturally avoids overfitting \cite{heckerman2008tutorial} and it is less sensitive to noise, because of its non-supervised nature and because it relies on a statistically based definition of conditional probability, instead of finding an arbitrary complex pattern fitting the data;
    \item it has shown good performances also with small sized dataset \cite{kontkanen1997comparing}, a common situation in medical applications.
\end{itemize}
On a less rigorous note, it has been observed that building causal graphs with BSL forces the researchers to think clearly about the topic, and articulate that thinking in the form of the graph, which is often beneficial in and of itself \cite{marcot2001using,walters2004fisheries,uusitalo2007advantages}.

All these features make BSL an optimal tool for medical research. In fact, it has found good application in the analysis of genetic data \cite{imoto2003bayesian,jiang2011learning}, electronic health record time series \cite{van2014learning}, longitudinal and standard clinical data \cite{onisko2016interpret,flores2011incorporating}.  In particular, the knowledge of causality relationship between potential predictors and mortality among patients diagnosed with COVID-19 would be crucial to target patients at highest risk and improve outcomes~\cite{potere2020acute}. However, only few works applied Bayesian approaches to the study of COVID-19, mainly to analyse its spreading \cite{dana2020brazilian,wibbens2020covid,mbuvha2020bayesian} and for contact tracing \cite{fenton2020privacy,mclachlan2020fundamental}. One paper interestingly applied BSL to serological tests of COVID-19 to determine a more accurate estimation of the disease's infection and fatality rates taking into account the reliability of each kind of test \cite{neil2020bayesian}.

Following these considerations, the main objective of this paper is to introduce the use of a BSL-based methodology to study the causal relationship between various predictors and mortality in COVID-19 patients. We aim at verifying whether BSL can reproduce and/or improve our current understanding of causal pathways of COVID-19 risk of mortality using a single-centre but feature-rich clinical dataset \cite{barbieri2020covid} which includes demographic information, clinical history of the patient, symptomatology, blood analysis data on admission and outcome. Our analysis is conducted both with and without incorporating prior medical information, to test whether there are significant differences.  Following up the causal analysis, we leverage the features that are deemed to be causally related to the outcome to construct a Binary Decision Tree (BDT) that describes a practical flow chart to identify patients at high-risk using data available right at the admission in hospital.

Overall, this study presents a novel clinical data analysis framework that provides a multi-step approach that goes from the identification of causal relationships between medical data to the realization of a clinically suitable tool for patients triage that is easy to use, straightforward to interpret and has solid clinical bases for its predictions.

\section{Dataset description} \label{sec:data}

The dataset includes COVID-19 diagnosed patients admitted between the 3rd of March 2020 and the 30th of April 2020 from three different units of the Pisa University Hospital (Emergency Room, Emergency Medicine Department and ICU)~\cite{barbieri2020covid}. All data were acquired from both paper and electronic records and carefully checked for the presence of spurious and/or erroneous inputs. Data collection was performed according to the principles stated in the Declaration of Helsinki and it conforms to standards currently applied in our country. The use of the data was approved by the Comitato Etico Area Vasta Nord Ovest (Internal Review Board - IRB number 230320). The patient’s informed consent was obtained.

The main outcome of the data cohort is dismissal at home or death. The subjects included in the study are 265, of whom 71 died for COVID-19 or for its complications. The death rate of this sample is thus 26.8\%. The original dataset included 125 variables. Among the various features collected during the triage and the hospitalization, 63 are missing in less than 50 subjects and thus are included in the study. For the purpose of this study, these features are grouped into six different logical categories as illustrated in Table \ref{tab:feature_summary}. All the medical exams used in this analysis have been made on the first day of hospitalization.


\begin{table}
\fontsize{7}{9}\selectfont
\centering
\begin{adjustwidth}{-2.25in}{0in}
\renewcommand{\arraystretch}{1.}
\begin{tabular}{|l|l|c|c|l|}
\hline
\multirow{3}{*}{Category} & 
\multirow{3}{*}{Feature} & 
\multirow{3}{*}{ \shortstack[l]{Available data \\for dead\\ subjects (max 71)} } & 
\multirow{3}{*}{\shortstack[l]{Available data \\for recovered \\ subjects (max 194)}} & 
\multirow{3}{*}{Values} \\
& & & & \\
& & & & \\
\hline

\multirow{4}{*}{ \shortstack[l]{Demographic \\ data} }& Age (\textcolor{blue}{years})   &   71   &   194  &  $\mu$ = 66.6 \;\; $\sigma$ = 15.9 \\
& SEX   &   71   &   194  &  0=male \;\; 1=female\\
& Smoke(y/n)   &   66   &   187  &  0=no \;\; 1=yes\\
& Smoke(ex/y/n)   &   64   &   187  &  0=yes \;\; 1=ex \;\; 2=no\\
\hline
\multirow{3}{*}{ \shortstack[l]{Prior \\ respiratory \\ problems}} & COPD (chronic obstructive pulmonary disease)  &   71   &   194  &  0=no \;\; 1=yes\\
& Asma   &   70   &   194  &  0=no \;\; 1=yes\\
& Other resp. disease   &   71   &   194  &  0=no \;\; 1=yes\\
\hline
\multirow{13}{*}{ \shortstack[l]{Prior \\ diseases}} & Diabetes   &   71   &   194  &  0=no \;\; 1=yes\\
& Hypertension   &   71   &   194  &  0=no \;\; 1=yes\\
& Cardio.disease   &   71   &   194  &  0=no \;\; 1=yes\\
& Hypercolest   &   71   &   194  &  0=no \;\; 1=yes\\
& Cerebrovasc. disease   &   71   &   194  &  0=no \;\; 1=yes\\
& Neuro. disease   &   71   &   194  &  0=no \;\; 1=yes\\
& Dementia   &   71   &   194  &  0=no \;\; 1=yes\\
& Cancer   &   71   &   194  &  0=no \;\; 1=yes\\
& Blood cancer   &   71   &   194  &  0=no \;\; 1=yes\\
& Kidney disease   &   71   &   194  &  0=no \;\; 1=yes\\
& Liver disease   &   71   &   194  &  0=no \;\; 1=yes\\
& Cirrhosis   &   71   &   194  &  0=no \;\; 1=yes\\
& Autoimmune disease   &   71   &   194  &  0=no \;\; 1=yes\\
\hline
\multirow{4}{*}{\shortstack[l]{Ongoing \\ treatments}} & Anticoag   &   71   &   194  &  0=no \;\; 1=yes\\
& RAAS BLOCK (renin-angiotensin-aldosterone system)   &   71   &   194  &  0=no \;\; 1=yes\\
& Immunos. therapy   &   71   &   194  &  0=no \;\; 1=yes\\
& Dialysis   &   71   &   194  &  0=no \;\; 1=yes\\
\hline
\multirow{19}{*}{\shortstack[l]{Symptoms \\on admission}} & Fever   &   71   &   194  &  0=no \;\; 1=yes\\
& Conjunct. congest.   &   71   &   194  &  0=no \;\; 1=yes\\
& Nasal congestion   &   71   &   194  &  0=no \;\; 1=yes\\
& Headache   &   71   &   194  &  0=no \;\; 1=yes\\
& Cough   &   71   &   194  &  0=no \;\; 1=yes\\
& Sore throat   &   71   &   194  &  0=no \;\; 1=yes\\
& Sputum   &   71   &   194  &  0=no \;\; 1=yes\\
& Fatigue   &   71   &   194  &  0=no \;\; 1=yes\\
& Hemoptysis   &   71   &   194  &  0=no \;\; 1=yes\\
& Short breath   &   71   &   194  &  0=no \;\; 1=yes\\
& Nausea   &   71   &   194  &  0=no \;\; 1=yes\\
& Diarrhea   &   71   &   194  &  0=no \;\; 1=yes\\
& Myalgia   &   71   &   194  &  0=no \;\; 1=yes\\
& Rash   &   71   &   194  &  0=no \;\; 1=yes\\
& FC   (cardiac frequency, \textcolor{blue}{bpm}) &   63   &   176  &  $\mu$ = 87.3 \;\; $\sigma$ = 17.8  \\
& PAS  (systolic arterial pressure, \textcolor{blue}{mmHg}) &   65   &   174  &  $\mu$ = 131.4 \;\; $\sigma$ = 20.3 \\
& PAD  (diastolic arterial pressure \textcolor{blue}{mmHg}) &   65   &   174  &  $\mu$ = 75.9 \;\; $\sigma$ = 13.3 \\
& Chest pain   &   71   &   194  &  0=no \;\; 1=yes\\
& Confusion   &   71   &   194  &  0=no \;\; 1=yes\\
\hline
\multirow{19}{*}{\shortstack[l]{Blood analysis\\ on admission}} & Haemoglobin (\textcolor{blue}{g/dl})   &   68   &   188  &   $\mu$ = 13.3 \;\; $\sigma$ = 2.01 \\
& WBC (white blood cells, \textcolor{blue}{cells/$n$l})  &   69   &   192  &  $\mu$ = 7.7 \;\; $\sigma$ = 3.8 \\
& Lymphocyte  (\textcolor{blue}{cells/$\mu$l}) &   68   &   191  &  $\mu$ = 1200 \;\; $\sigma$ = 1140 \\
& Neutrophils (\textcolor{blue}{cells/$\mu$l})  &   69   &   188  &  $\mu$ = 5690 \;\; $\sigma$ = 3322 \\
& Haematocrit  (\textcolor{blue}{$\%$})  &   69   &   187  &  $\mu$ = 39.7 \;\; $\sigma$ = 22.2 \\
& Platelets  (\textcolor{blue}{cells/$n$l}) &   69   &   191  &  $\mu$ = 206.3 \;\; $\sigma$ = 102.1\\
& INR (international normalized ratio) &   65   &   184  &  $\mu$ = 1.81 \;\; $\sigma$ = 7.38 \\
& Bilirubin (\textcolor{blue}{mg/dl})  &   64   &   186  &  $\mu$ = 1.26 \;\; $\sigma$ = 6.13 \\
& AST (aspartate aminotransferase, \textcolor{blue}{IU/l})  &   59   &   181  &  $\mu$ = 44.70 \;\; $\sigma$ = 46.18 \\
& ALT (alanine aminotransferase, \textcolor{blue}{IU/l}) &   66   &   187  &  $\mu$ = 42.77 \;\; $\sigma$ = 49.81 \\
& Glucose  (\textcolor{blue}{mg/dl}) &   64   &   186  &  $\mu$ = 127.4\;\; $\sigma$ = 46.3\\
& Creatinine (\textcolor{blue}{mg/dl}) &   67   &   191  &  $\mu$ = 1.22 \;\; $\sigma$ = 1.09\\
& BUN (blood urea nitrogen,\textcolor{blue}{mg/dl})  &   58   &   178  &  $\mu$ = 27.9 \;\; $\sigma$ = 24.8 \\
& Sodium (\textcolor{blue}{mEq/l})  &   68   &   190  &  $\mu$ = 138.1 \;\; $\sigma$ = 4.6 \\
& Potassium (\textcolor{blue}{mmol/l})  &   67   &   187  &  $\mu$ = 4.02 \;\; $\sigma$ = 0.67 \\
& pH   &   57   &   162  &  $\mu$ = 7.45 \;\; $\sigma$ = 0.06 \\
& pO2 (O2 partial pressure, \textcolor{blue}{mmHg})  &   62   &   173  &  $\mu$ = 75.3\;\; $\sigma$ = 32.7 \\
& pCO2 (CO2 partial pressure, \textcolor{blue}{mmHg})  &   61   &   167  &  $\mu$ = 34.6 \;\; $\sigma$ = 7.9 \\
& PF (pO2/FIO2 ratio,  \textcolor{blue}{$\%$})   &   65   &   188  &  $\mu$ = 283.2 \;\; $\sigma$ = 95.8 \\
& PCR (C-reactive protein ,\textcolor{blue}{mg/dl})   &   67   &   177  &  $\mu$ = 9.25\;\; $\sigma$ = 8.55 \\
\hline
\multicolumn{2}{|c|}{\multirow{2}{*}{Outcome}}  &   \multirow{2}{*}{71}   &   \multirow{2}{*}{194}  &  \multirow{2}{*}{\shortstack[l]{0=death \\ 1=recovery}}\\
\multicolumn{2}{|c|}{} & & & \\

\hline

\end{tabular}
\caption{Summary of the dataset features. Third and fourth column show the feature occurrence in the dataset in dead and recovered subjects, while the last column reports a description of the feature values.} 
\label{tab:feature_summary}
\end{adjustwidth}
\end{table}

\section{Causal learning for COVID-19 data} \label{sec:method}

The methodology put forward in this paper is articulated in three steps, leveraging different machine learning and statistical methods. First we apply BSL to analyze how the variables of the sample are causally interconnected. Then, we evaluate with Bivariate Statistical Tests (BST) the strength of the connections between the outcome and its neighbours in the BSL graph. This second analysis may seem redundant with respect to the first step. However, while BSL performs a multivariate analysis taking into consideration also latent variables and the possibility of mediated connections, the second one evaluates singularly how each variable affects the outcome. Thus, on the one hand, the causal graph obtained with BSL helps to understand the logical cause-effect sequence bringing to the outcome and to select the features of interest; on the other hand, BST helps to rank the dependency between each feature and the outcome. The third step targets selecting the most relevant features to train a Binary Decision Tree (BDT) useful for clinical practice. 

The following sections provide a synthetic background on Bayesian Networks, describe the BSL algorithm used and the workflow of the three analyses discussed above. 

\subsection{Bayesian Network (BNs) background}
BNs are graphical models ${(G,P)}$, where $P$ is a joint probability of random variables
${X_{V}=(X_{1},\dots,X_{N})}$ associated with nodes ${V}=\{1,\dots,N\}$. Each random variable $X_i$ can possibly be of different nature (binomial, multinomial, ordinal, continuous).  The graph ${G=(V,E)}$ is a directed acyclic graph (DAG) whose edges ${E}$ encode the joint probabilistic relationships among the $N$ random variables. The graph is a visual representation of the joint distribution of the data, where a directed edge ${e_{ij}}$ from node ${i}$ to ${j}$ indicates that ${i}$ is the {\it parent} of ${j}$ as part of a conditional
dependence relationship between the two nodes \cite{beretta2018learning}.

Broadly speaking, there are two main families of methods to learn the structure of a BN from data: constraint-based methods and score-based ones \cite{buntine1996guide,daly2011learning}. The first class of methods learns the conditional independence relations of the BN from which, in turn, it generates the network. Score-based approaches, instead, cast structure learning as an optimization problem, often addressed as an heuristic search task leveraging a score function to drive exploration of the space of the graph structures in search of the optimal DAG.

\subsection{Causal graph analysis}

\subsubsection{BSL Algorithm}
In order to single out the best approach for our analysis, we have considered the fact that constraint-based algorithms have been shown to be more accurate than score-based algorithms for small sample sizes, and that score based algorithms tend to scale less well to high-dimensional data \cite{scutari2019learns}. Additionally, constraint-based algorithms can naturally integrate domain knowledge in the form of already-known results of independence tests, while it is more difficult to include this kind of information in score-based methods. 
Since we would like to incorporate prior knowledge in our analysis, and since the single samples in our dataset are highly dimensional while the dataset has a relatively low number of subjects, in this paper we use a constraint-based algorithm called Fast Causal Inference (FCI)\footnote{Implemented using the \textit{fci} function of the {\tt pcalg} R package} \cite{spirtes2000causation}. FCI is an extension of the popular PC \cite{spirtes1991algorithm} algorithm that considers also the presence of hidden variables, that are random variables for which no observable input data is available. Briefly, the FCI algorithm comprises two steps:
\begin{itemize}
    \item Skeleton definition: starting from a fully connected graph, edges are removed according to the result of conditional independence tests between pairs of variables, or to the domain knowledge provided;
    \item DAG learning: the edge directions are learnt and added to the skeleton.
\end{itemize}
In our analysis, variable independence is tested with a Gaussian conditional independence test for continuous variables, and the $G^2$ test for binary variables \cite{spirtes2000causation}.

A significant issue with constraint-based algorithms is that the number of independence tests needed to remove an edge increases significantly with the number of vertices and with the number of values assumed by the random variables (for the discrete case). For this reason, heuristic search algorithms are usually employed to avoid testing a combinatorial number of cases \cite{spirtes2000causation}. As a consequence, the true DAG can seldom be identified and multiple DAGs with different edge directions are found to be compatible with the input data. For this reason, in our analysis we do not consider edge orientation by the algorithm as reliable information, leaving this task to the clinician interpretation.

\subsubsection{Quantifying causal effects}\label{thickness}

For every DAG identified by FCI, we estimate the effect that each variable has on the variables it is connected to, representing an indicator of the strength of their causal connection, using covariate adjustment\footnote{Implemented using the \textit{ida} function of the {\tt pcalg} R package}. 
Basically, for each pair of variables, $X$ and $Y$, we compute a linear regression of $Y$ on $X$ using the parent set of $X$, extracted from the DAG, as covariate adjustment.
When multiple DAGs are available, we average the effects estimated from each DAG. However, in this study we have observed that the DAG is usually estimated without ambiguity and thus the cases in which multiple different effects are found are very rare.
Note that this method describes the effect that a unitary increase in $X$ has on $Y$ and, therefore, it is highly sensitive to the range of both $X$ and $Y$. In order to be able to qualitatively compare the effects, we standardized the variables before this calculation.

\subsubsection{BSL strategy on our cohort}

The application of the BSL pipeline described in the previous subsection followed some considerations related to the nature of the dataset available in our study. First, it has to be noted that only a portion of the features in Table \ref{tab:feature_summary} has been collected or is available for all subjects in our dataset. As most data-driven methods, BSL performs better when the sample size is higher than the number of features. Therefore, we opted for 
applying the BSL pipeline separately for the 6 categories of features identified in Table \ref{tab:feature_summary}, in order to maximize the amount of samples available. The outcome variable has been included in each category-specific analysis. 

After building the BNs (one per category), we identified those features of each skeleton that reported either a direct causal connection with the outcome, or an indirect connection mediated by a single intermediate feature. These served to define the candidate pool of relevant features which have been jointly analyzed with another round of the BSL algorithm (i.e. representing samples only in terms of the filtered features) leading to an integrated causal graph.

The BSL algorithm attempts to infer causal relationships from input data, which are unavoidably biased, because the sample analyzed is composed by subjects whose medical conditions required hospitalization. Thus, the results from this analysis should be interpreted as representative of the most severe situations. 

The causal graph analysis pipeline discussed so far has been repeated under two different conditions. 
In the first we provide to the BSL process a conservative list of  prohibited causal connections (drawn up by our clinicians and based on the fact that there is vast medical agreement on the nonexistence of such relationships), whose severance is enforced during the graph identification process. The second one, instead, comprises a fully data driven setting, where no prior knowledge is supplied to the BSL algorithm.
This second condition is tested for comparison, to assess the capability of the method to autonomously identify causal relationships that are not blatantly in contrast with widely agreed clinical knowledge.

\subsection{Bivariate statistical analysis}
The focus of this step of our analysis is to study how the outcome depends singularly from each relevant feature identified at the previous stage. To this end, we use the Fisher Exact Test (FET) \cite{finney1948fisher} for categorical variables and the Point-Biserial Correlation (PBC) \cite{kornbrot2005point} for the continuous ones. The FET method tests whether two binary variables are independent, such as "death-recovery" and "male-female". The PBC, instead, is a correlation test, such as the Pearson's correlation, specifically developed to correctly estimate the correlation between a categorical variable, in this case the outcome, and a continuous variable. The bivariate statistical analysis is applied to all the features considered in the integrated causal graph.

\subsection{Decision tree analysis}
BDT models are characterized by an inferential process which has some resemblance with human reasoning, which makes them amenable to human interpretation. However, they are known to severely suffer from overfitting \cite{bramer2007avoiding}, thus it is fundamental to reduce the complexity, and thus the depth, of the tree to avoid the identification of poorly generalizable classification rules. 

The scope of our BDT analysis is two-fold. On the one hand we would like to identify a compact BDT to provide an interpretable and procedural view on what are the discriminative cut-point values for the features identified by the BSL causal analysis, and if those values are in agreement with clinical knowledge. On the other hand, we would like to measure the predictive value of such candidate features on our dataset. 

The first objective (interpretation) has been addressed by training a BDT with a depth fixed to 4 levels, considering only those features present in the integrated graph and only those subjects having those features fully available. The interpretable BDT is trained on the full data, as it is never used for prediction.  To assess the impact of the identified causal features on predictive performance, we train two additional models. The first is a BDT with the same structural constraints and features of the intepretable one, but it is trained and assessed in a 10-fold cross-validation scheme, Its predictive performance is assessed as the average accuracy on the 10 validation folds. The second is a permutation test BDT, again trained and evaluated in 10-fold cross-validation, but using features randomly drawn from the set of features in Table \ref{tab:feature_summary}.

The comparison between the latter two BDTs is intended to verify the effectiveness of the feature identified by the BSL causal analysis. However, a fair comparison requires to ensure that the two BTDs have a comparable number of subjects in training, otherwise it is impossible to understand whether a performance reduction is to be imputed to the lower training size or to the different features. Thus, in the permutation test, we train about $1000$ different BDTs of depth-4 with number of training subjects and features matching those used for training the causal BDT.

\section{Results}

\begin{figure}
\centering
  \includegraphics[width=.9\textwidth]{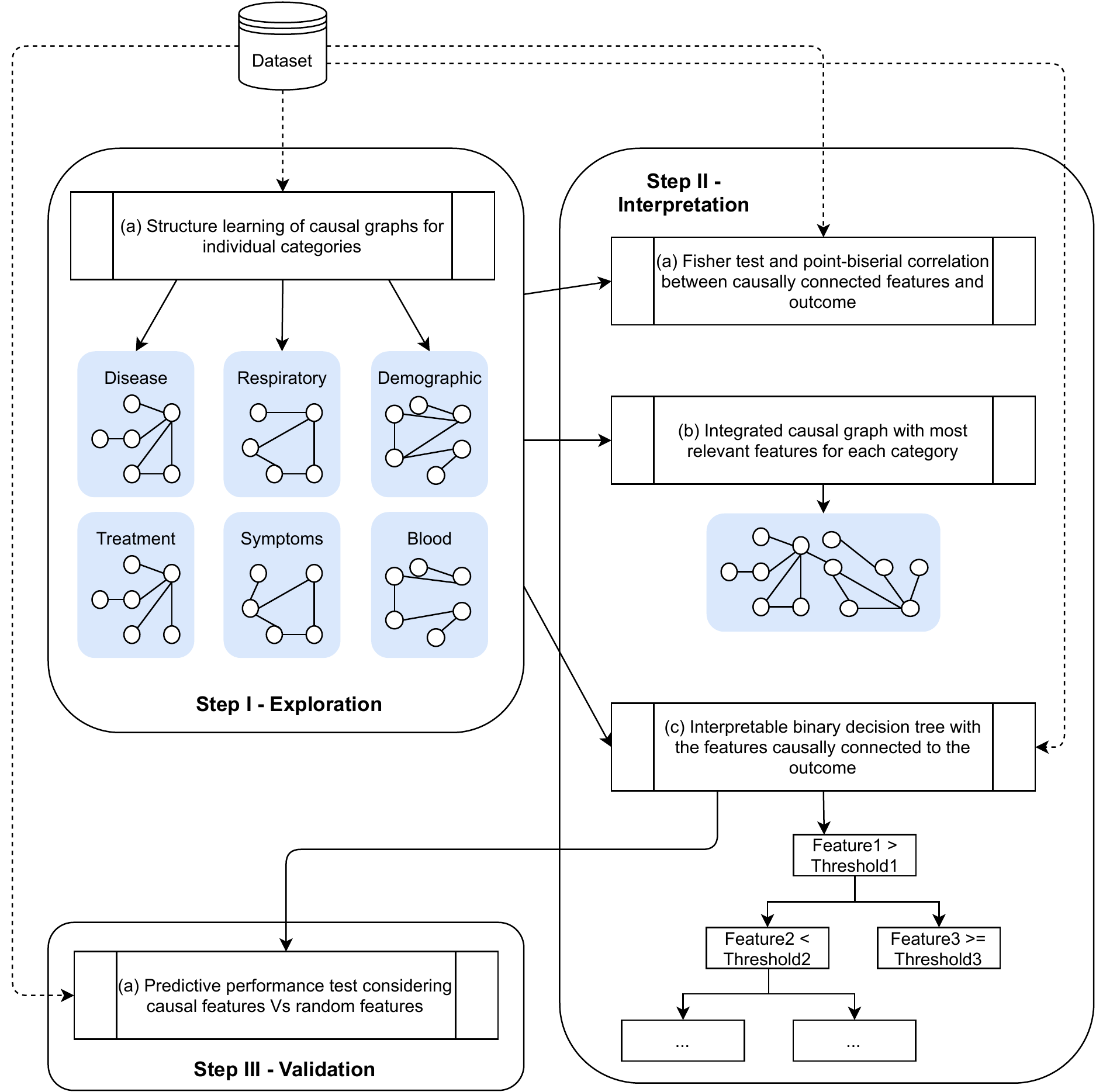}
  \caption{Flow chart of the empirical analysis highlighting the three steps, from explorative analysis, to interpretation, concluding with quantitative validation. \label{fig:flowchart}}
\end{figure}

We report the main results of the empirical analysis run on the dataset described in Section \ref{sec:data} following the three-fold structure of our methodology (Section \ref{sec:method}). A pictorial overview of the experimental analysis is provided in Figure \ref{fig:flowchart}. Here, we highlight the explorative causal analysis run separately on the different feature categories (Step I), followed by the  analysis on selected causal features (Step II) comprising the integrated causal graph, the BST and interpretable BDT. The final stage comprises a predictive analysis comparing the BDT on causal features with the BDT on randomly drawn features (Step III).

\subsection{Causal graph analysis}
\begin{figure}
\begin{adjustwidth}{-2.25in}{0in}
\centering
  \includegraphics[width=1.1\textwidth]{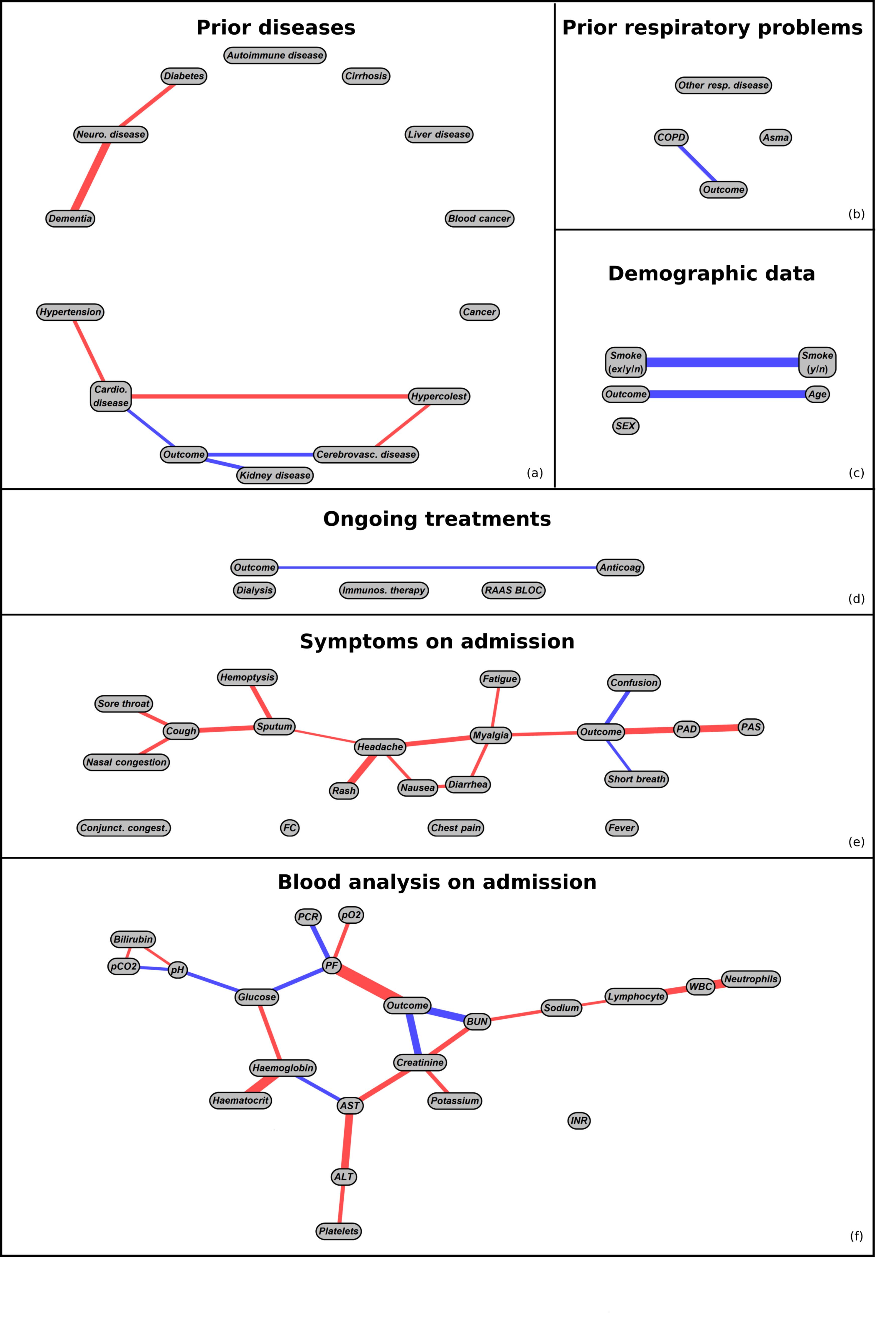}
  \caption{Image showing the BSL analysis applied to different categories of features. All the illustrated graphs are generated taking the information provided by clinicians into account.}
  \label{fig_grafi_con_info}
  \end{adjustwidth}
\end{figure}

\begin{figure}
\begin{adjustwidth}{-2.25in}{0in}
\centering
  \includegraphics[width=1.1\textwidth]{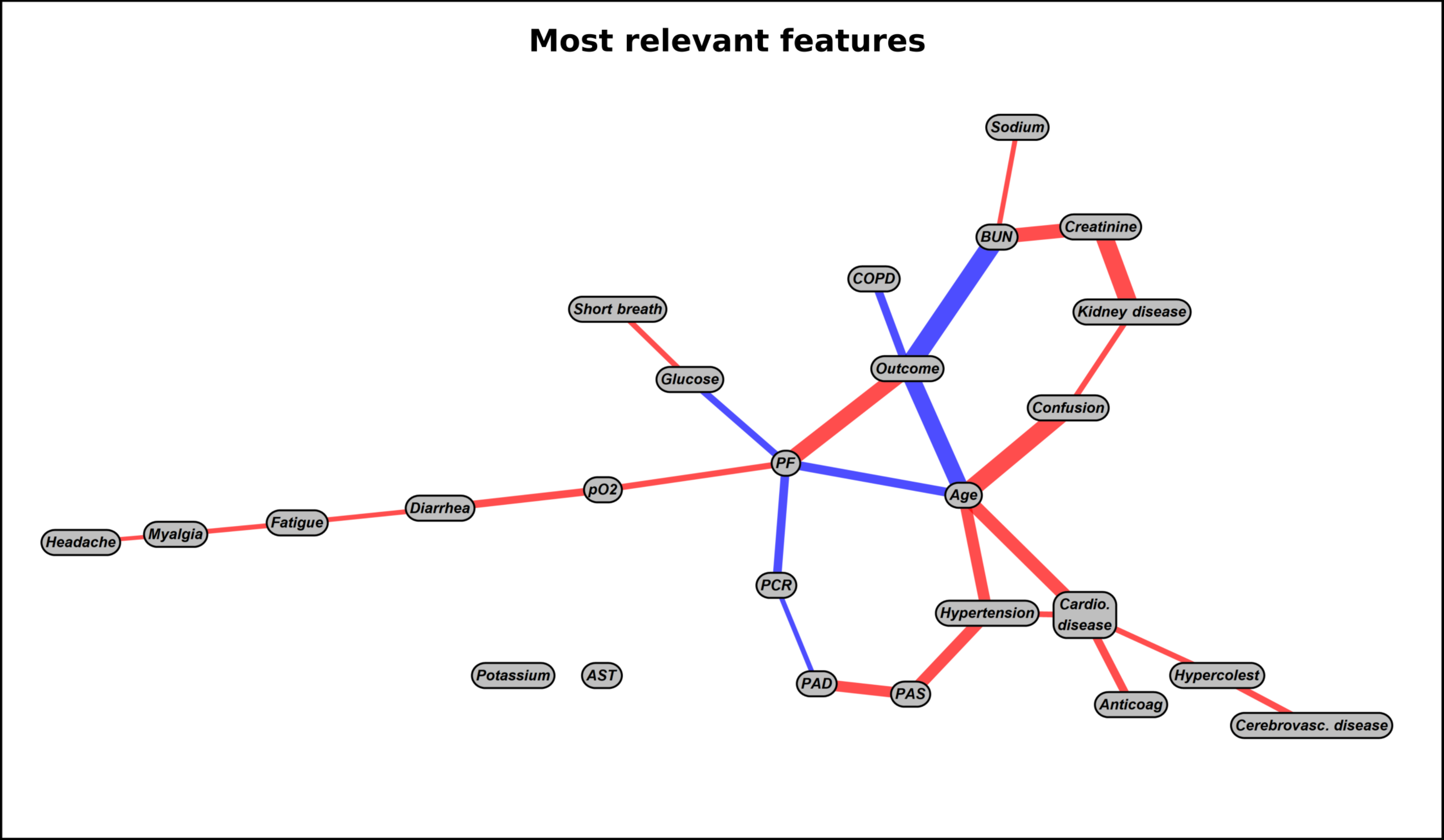}
  \caption{Graph generated with the most relevant features found from the graphs shown in Fig. \ref{fig_grafi_con_info}, taking the information provided by clinicians into account}
  \label{fig_grafo_globale_info}
  \end{adjustwidth}
\end{figure}

The results of the BSL step run separately on the single feature categories are provided in Figure \ref{fig_grafi_con_info}, while Figure \ref{fig_grafo_globale_info} depicts the corresponding integrated graph.  All graphs have been obtained including prior knowledge on prohibited connections through a list provided by our physicians. Their fully data-driven counterparts are reported in 
the Appendix (Figure \ref{fig_grafi_no_info} and \ref{fig_grafo_globale_no_info}) to ease readability. One can easily appreciate that no substantial difference exists between the two sets of graphs, confirming the robustness of the causal inference process. Note that the distance between nodes in the graphs has no meaning and it is automatically adjusted to make the diagrams visually appealing. Instead, the thickness of the edges reflects the intensity of the connection (calculated as described in Section \ref{thickness}) and the red (blue) edges connect positively (negatively) correlated variables. The definition of positively/negatively correlated variables may be confusing when one variable is categorical and the other is continuous. For instance in Figure \ref{fig_grafo_globale_info}, age (continuous) appears to be negatively correlated with the outcome, which has value $1$ for death and $0$ for recovery. This means that older subjects have a higher risk of death than younger ones. To fully understand the meaning of each connection, we refer to Table \ref{tab:feature_summary} which reports the values of each categorical variable under examination.

\subsection{Bivariate statistical analysis}
Table \ref{tab:contingency} reports the contingency and the p-values of the Fisher exact test computed for all the categorical variables found relevant in the single-category BSL analyses and thus included in the integrated causal graph in Figure \ref{fig_grafo_globale_info}.
With \textit{relevant} we mean that these features are connected to the outcome within 2 hops.
To ease intepretation of the results we report in red (green) colour the death (recovery) excess factors with respect to the occurrence in the whole dataset. 

Table \ref{tab:pbc} instead reports PBCs and their p-values. As it can be noted, these measures emphasize some differences with respect to the BSL-based analysis. For instance, in the integrated graph fatigue and headache seem both to be linked to the outcome through myalgia. The Fisher analysis instead reveals that headache is unrelated to the outcome and has to be simply related to myalgia, while fatigue seems to be actually causally connected to the outcome.

\definecolor{officegreen}{rgb}{0.0, 0.5, 0.0}

\begin{table}
\begin{adjustwidth}{-2.25in}{0in}
\centering
\renewcommand{\arraystretch}{1.1} 
\small
\begin{tabular}{|l|r@{}l|r@{}l|c|c|c|c|}
\hline
\multirow{2}{*}{Feature X} & 
\multicolumn{2}{c|}{\multirow{2}{*}{ \shortstack{\% of deaths in \\ patients with X} }} & 
\multicolumn{2}{c|}{\multirow{2}{*}{ \shortstack{\% of recoveries in \\ patients with X} }} & 
\multirow{2}{*}{\shortstack{\% of deaths in \\ patients without X}} & 
\multirow{2}{*}{\shortstack{\% of recoveries in \\ patients without X}} & 
\multirow{2}{*}{\shortstack{P-Value \\ (Fisher test)}}\\
& & & & & & & \\
\hline
\hline
COPD &  \hspace{2mm} 69.0\hspace{1mm}\% & \hspace{1mm} $ {\color{red} (\times 2.6)} $ & 31.0 \% & & 21.6 \% & 78.4 \% & $5.2*10^{-7}$   \\
Kidney disease & 58.3\hspace{1mm}\% & \hspace{1mm} ${\color{red} (\times 2.2)} $ & 41.7 \% & & 23.7 \% & 76.3 \% & $6.2*10^{-4}$   \\
Cerebrovasc. disease &  59.3\hspace{1mm}\% & \hspace{1mm} ${\color{red} (\times 2.2)} $ & \hspace{2mm} 40.7 \% &  & 23.1 \% & 76.9 \% & $1.7*10^{-4}$   \\
Cardio. disease &  42.5\hspace{1mm}\% & \hspace{1mm} ${\color{red} (\times 1.6)}$ & 57.5 \% & & 19.1 \% & 80.9 \% & $6.4*10^{-5}$   \\
Anticoag. &  51.5\hspace{1mm}\% & \hspace{1mm} ${\color{red} (\times 1.9)} $ & 48.5 \% & & 23.3 \% & 76.7 \% & $1.1*10^{-3}$   \\
Myalgia & 2.7 \% & & 97.3\hspace{1mm}\% & \hspace{1mm} ${\color{officegreen} (\times 1.3)}$ & 30.7 \% & 69.3 \% & $6.1*10^{-5}$   \\
Confusion &  71.9\hspace{1mm}\% & \hspace{1mm} ${\color{red} (\times 2.7)}$ & 28.1 \% & & 20.6 \% & 79.4 \% & $1.4*10^{-8}$   \\
Short breath &  34.1\hspace{1mm}\% & \hspace{1mm} ${\color{red} (\times 1.3) }$ & 65.9 \% & & 20.1 \% & 79.9 \% & $7.5*10^{-3}$   \\
Dialysis &  50.0\hspace{1mm}\% & \hspace{1mm} ${\color{red} (\times 1.9) }$ & 50.0 \% & & 26.4 \% & 73.6 \% & $0.29$   \\
Hypercolesterolemia & 37.0\hspace{1mm}\% & \hspace{1mm} ${\color{red} (\times 1.4) }$ & 63.0 \% & & 24.7 \% & 75.3 \% & $6.6 * 10^{-2}$   \\
Hypertension &  35.5\hspace{1mm}\% & \hspace{1mm} ${\color{red} (\times 1.3)} $ & 64.5 \% & & 19.1 \% & 80.9 \% & $2.1 * 10^{-3}$   \\
Diarrhea & 11.8 \% & & 88.2\hspace{1mm}\% & \hspace{1mm} ${\color{officegreen} (\times 1.2)} $ & 30.4 \% & 69.6 \% & $4.0 * 10^{-3}$   \\
Fatigue  & 23.3 \% & & 76.7\hspace{1mm}\% & \hspace{1mm}  & 27.5 \% & 72.5 \% & $0.36$   \\
Headache &  4.3 \% & & 95.7\hspace{1mm}\% & \hspace{1mm} ${\color{officegreen} (\times 1.3)} $ & 28.9 \% & 71.1 \% & $5.5 * 10^{-3}$   \\
\hline
\end{tabular}
\caption{Contingency table and results of the Fisher test between the categorical variables present in Fig. \ref{fig_grafo_globale_info} and the outcome. The deaths (recoveries) fold increase with respect to the dataset average is marked in red (green).}
\label{tab:contingency}
\end{adjustwidth}
\end{table}

\begin{table}
\centering
\renewcommand{\arraystretch}{1.3} 
\begin{tabular}{|l|r|c|}
\hline
Feature & Correlation & P-Value \\
\hline

Age             &  -0.46    &   $1.3*10^{-15}$ \\
PAS             &   0.21    &   $4.6*10^{-4}$  \\
PAD             &   0.24    &   $8.1*10^{-5}$  \\
AST             &  -0.11    &   0.052  \\
Glucose         &  -0.19    &   $1.5*10^{-3}$  \\
Creatinine      &  -0.20    &   $5.6*10^{-4}$  \\
BUN             &  -0.45    &   $1.1*10^{-13}$ \\
Sodium          &  -0.16    &   $5.0*10^{-3}$  \\
Potassium       &  -0.18    &   $1.6*10^{-3}$  \\
PCR             &  -0.21    &   $4.4*10^{-4}$  \\
pO2             &   0.12    &   0.032 \\
PF              &   0.46    &   $1.1*10^{-14}$\\

\hline

\end{tabular}
\caption{Point-biserial correlation value and significance test between the continuous variables present in Fig. \ref{fig_grafo_globale_info} and the outcome.} 
\label{tab:pbc}
\end{table}

\subsection{Decision tree analysis}
\begin{figure}
\begin{adjustwidth}{-2.25in}{0in}
\centering
  \includegraphics[width=\textwidth]{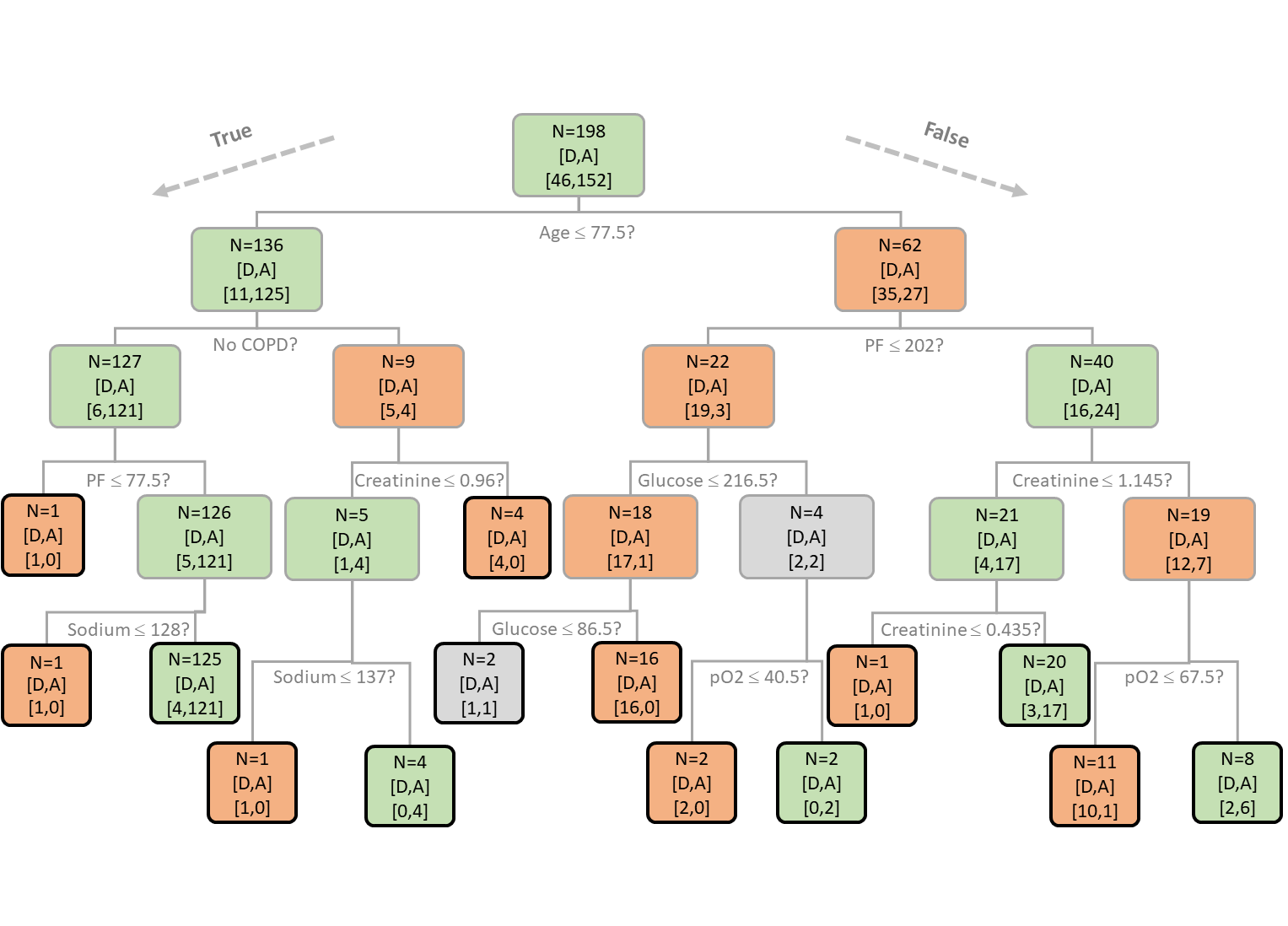}
  \caption{BDT trained with the features included in the graph analysis reported in Fig. \ref{fig_grafo_globale_info}. The color of the squares represents the class prevalence: red for deaths, green for recoveries and grey in case of parity. The number of subjects is indicated in the first line of each square, while the last line reports deaths/recoveries. Black edges denote leaves.  }
  \label{fig:tree}
  \end{adjustwidth}
\end{figure}

Figure \ref{fig:tree} shows the BDT trained with all  the 198 subjects having the 26 features included in the integrated causal graph Fig. \ref{fig_grafo_globale_info}. The interpretable BDT chose to base its prediction exclusively on $7$ features: AGE, PF, COPD, Creatinine, Glucose, pO2 and Sodium. Remarkably, all of them are connected to the outcome within 2 hops in the integrated causal graph (Figure \ref{fig_grafo_globale_info}). When computing the classification accuracy of the interpretable BDT on its training set, we appreciate that only 5.5\% of the data are misclassified. Despite a few leaves being clearly tailored to fit the dataset (and thus being good candidates for pruning), all the learned patterns appear to be in line with current clinical understanding of how these features affects the outcome.

The first line in Table \ref{tab:performance} shows various performance metrics of the BDT classifier evaluated in "train all" setting (i.e. trained and validated on the same data comprising all the available subjects) and in a 10 fold cross-validation assessment. As it can be noted, the results are quite coherent, suggesting that the overall pattern is really informative and almost insensitive to overfitting.
The permutation test, whose results are reported in the second line of Table \ref{tab:performance}, shows instead a completely different behaviour. Even with randomly selected features it is possible to achieve high performances in "train all" setting, but these performances are not maintained in cross-validation.

\begin{figure}
\begin{adjustwidth}{-2.25in}{0in}
\centering
  \includegraphics[width=0.7\textwidth]{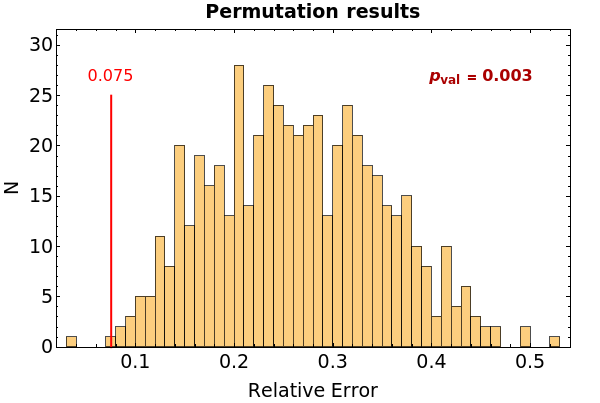}
  \caption{Histogram reporting the percentage of misclassified subjects in the permutation test. The red bar shows the performance of the BDT reported in Fig. \ref{fig:tree}. All the misclassification rates are calculated as the average over a 10-fold cross-validation.}
  \label{fig:histperm}
  \end{adjustwidth}
\end{figure}

Figure \ref{fig:histperm} depicts the average proportion of misclassified patients in cross-validation of our BDT (red line) and of the permutation test (orange columns), showing that the errors in the former tree are significantly lower from what would be obtained from one considering 7 features randomly taken from the dataset.

We believe that these results support the methodological choice to select the features causally connected to the outcome through BSL and then training a BDT to obtain an intepretable clinical decisional tool.

\begin{table}
\centering
\begin{adjustwidth}{-2.25in}{0in}
\renewcommand{\arraystretch}{1.3} 
\begin{tabular}{|l|c|c|c|c|c|c|}
\hline
\multirow{2}{*}{Input data} & \multicolumn{2}{|c|}{Sensitivity} & \multicolumn{2}{|c|}{Specificity} & \multicolumn{2}{|c|}{F1 score} \\\cline{2-7}
& Train all & 10f cross-val  &  Train all & 10f cross-val & Train all & 10f cross-val\\
\hline
\hline

7 features of the tree in Fig. \ref{fig:tree}& 0.99 & 0.90 & 0.95 & 0.95 & 0.97 & 0.92\\
7 random features &0.99 &0.88& 0.82&0.49& 0.97& 0.86\\

\hline

\end{tabular}
\caption{Comparison between the performance of the classifier developed in this study and classifiers trained on a random set of 7 features. The results of the permutation tests are the average of those obtained from all the 1,000 permutations.} 
\label{tab:performance}
\end{adjustwidth}
\end{table}

\section{Discussion}
In this section, we briefly discuss the experimental results obtained by our causal analysis, from the clinical perspective.
We begin by considering the results from the single-category causal graphs in Figure \ref{fig_grafi_con_info}. 

Among respiratory diseases (graph b), COPD is the only pathology in direct causal relationship with mortality. This result was recently confirmed by a meta-analysis on the associations between respiratory diseases and COVID-19 outcomes~\cite{sanchez2020underlying}. 

As regards prior (non-respiratory) diseases (graph a), not surprisingly, cardiovascular disease has a direct relationship with the outcome~\cite{gacche2021predictors}. Furthermore, it is the mediator of the causal effects of hypertension and hypercholesterolemia. On the other hand, hypercholesterolemia also has a causal effect on the presence of cerebrovascular disease which has a direct impact on the outcome~\cite{xu2020association}. Similarly, the presence of dialysis has only an indirect causal relationship on mortality through kidney disease~\cite{fathi2021prognostic}. Remarkably, the three pathologies directly connected to the outcome (i.e. cardiovascular, cerebrovascular and kidney diseases) exert a strong effect on it, increasing the probability of death by $1.6$, $2.2$ and $2.2$ with respect to the average in the dataset, respectively (see the analysis in Table \ref{tab:contingency}).
In contrast to published results, in our analysis liver diseases (including cirrhosis)~\cite{sharma2020liver}, diabetes~\cite{coppelli2020hyperglycemia} and the presence of dementia or neurological diseases ~\cite{july2021prevalence} seem to be not causally connected to the outcome. This discrepancy is mainly imputable to their scarce representation in the dataset, which does not allow to satisfy the conditional probability criterion with which the graph connections are identified.
In fact, despite their limited size, in all the subsets of subjects presenting the aforementioned diseases, the risk of mortality rises at least by a factor of 2.3. However, these increases cannot be attributed to these diseases alone because at least 75\% of the subjects who died and were affected by one of them, presented also one of the three pathologies directly connected to the outcome.

Age is one of the most cited risk factors due to its close relationship with mortality in COVID-19 patients~\cite{yadaw2020clinical}. Accordingly, the BSL approach (see graph c) identifies a strong causal relationship with the outcome also in the patients enrolled in Pisa. The importance of this effect can be appreciated even more by looking at the thickness of the "age-outcome" edge in the integrated causal graph, Fig. \ref{fig_grafo_globale_info}.
On the other hand, in our cohort neither sex nor smoking seem to have a direct causal role with mortality contrary to what was claimed by recent meta-analysis~\cite{peckham2020male,zhang2021association}.
The apparent inconsistency on the effect of sex is explained by the intrinsically biased cohort which has been selected based on the symptomatology severity of the patients in admission. In fact, only 32\% of the examined sample is composed by females, which means that they are somewhat protected from acute forms of COVID-19. Additionally, it must be noted that in an unbiased situation hospitalized females should be even more than males, considering that they have a higher longevity \cite{austad2006women} and thus the elder population, which is the most vulnerable to COVID-19 is enriched by females. Thus, according to our analysis being a female has no protective effect against death when the patient already shows severe symptoms that necessitate hospitalization, while this protective effect is present \textit{ab origin} and it is detectable by the gender bias in hospitalized patients. 

The BSL analysis applied to treatment features (graph d) shows also a direct connection between the assumption of anticoagulant before/during Coronavirus infection and the outcome. As it can be noted from Table \ref{tab:contingency}, the prescription of anticoagulants increases the risk of mortality by a factor of 1.9, while the cardiovascular diseases by only 1.6. This effect suggests that the subjects that are prescribed to take anticoagulants have a more severe form of cardiovascular disease, and that the true causal effect is given by cardiovascular disease alone.
Following this idea, given that kidney diseases increase the risk of mortality, we would expect a connection between outcome and dialysis, however only 4 subjects of the cohort were following this treatment. 

Results from the analysis of the symptoms present at hospital admission (graph e) are particularly intriguing. In fact, only a relatively small number of symptoms show a direct causal relationship with the outcome. Among these, myalgia seems to be protective against the risk of mortality. This fact, which has been recently confirmed by Zheng et al.~\cite{zheng2020risk}, could be explained by the fact that patients with less severe form of disease are reporting a generic muscle pain with augmented sensitivity as they do not suffer from more severe and debilitating symptoms. Accordingly, myalgia presents upstream causal relationships with many other symptoms of medium and mild severity, such as nasal congestion, sputum and nausea.
On the other hand, a strong causal link between confusion and mortality was detected (leading to an increased risk of death by $2.7$ with respect to the average in Table \ref{tab:contingency}), probably associated with the older age of patients presenting this symptom.
Regarding shortness of breath, its association with outcome can be a consequence of the effects of COPD on mortality. In the same way, the effects of PAD (and indirectly PAS) could be explained by the already demonstrated causal relationship between cardiovascular disease and outcome.
In contrast to what is present in the literature, the presence of fever at the time of admission seems to lack a direct causality link with mortality~\cite{qiu2020clinical}, however differences in fever assessment may alter the results of this analysis. For instance, fever may be measured at the hospital at the time of admission or assessed by the patient and in both cases its value may be altered by antipyretic assumption, whose frequency depends on self-medication habits.

Among laboratory tests (graph f), only three parameters have a direct causal relationship with mortality. PF ratio is an index of respiratory function and, given that the respiratory system is one of the systems most affected by COVID-19, its causal relationship with mortality is highly understandable. This connection can be observed both in the single-category as well as in the integrated causal graph. Other clinical variables selected by our automatic method are BUN and creatinine. Elevated levels of both BUN and creatinine can be associated to chronic kidney disease which was already demonstrated in direct causal relationship with mortality. Furthermore, a link between creatinine, BUN and mortality has already been observed in COVID-19 patients by other authors~\cite{cheng2020kidney}.

The study of causality between classes of clinical variables and mortality in COVID-19 patients revealed some already known relationships as well as offering some new insights. However, the picture becomes more clear when all clinical variables are used together to assess causal effects on the outcome as in the integrated causal graph of Figure \ref{fig_grafo_globale_info}. In this case, only 4 variables show a direct causal effect on the outcome. However, these direct relationships are very strong, with PF, age and BUN  being particularly correlated to mortality (see Table \ref{tab:pbc}).
Furthermore, these parameters also exhibit a strong upstream causal relationship with practically all the clinical variables that are directly associated with the outcome in the analysis divided by single categories. In fact, age has causal relationships with confusion, cardiovascular and cerebrovascular diseases, as well as with the use of anticoagulants and high blood pressure. Similarly, BUN levels mask a causal relationship with creatinine levels and the presence of nephrological disease. Finally, the PF mediates the relationship with shortness of breath. On the contrary, myalgia seems to have a weaker and more distant causal role with the outcome when all the clinical variables are considered. This seems to sustain the hypothesis that the apparently causal of myalgia on symptoms is due to the fact that less severe patients report its occurrence together with the deficiency of more critical symptoms.

Moving to the BDT illustrated in Fig. \ref{fig:tree}, remarkably the first two levels of the tree alone allow to correctly classify 85\% of the subjects basing on their age, a previous history of COPD in young and middle-aged patients and the PF ratio value in the older ones. The PF threshold discovered by the algorithm to distinguish patients at high and low risk is almost equal to the criterion adopted by the American-European Consensus Conference Committee (AECC) to define hypoxemia, i.e. a PF ratio less than or equal to 200 mmHg \cite{ware2000acute}. Throughout the tree, creatinine levels have been questioned 3 times. Despite in healthy subjects creatinine normally can range between 0.6 to 1.3 mg/dl, from this analysis it seems that subjects affected by COVID-19 have an increased risk of mortality already when its value passes 0.96 for the younger ones and 1.145 for the older ones. 
Instead, binary decisions based on sodium and pO2 levels are coherent with normal range values. Sodium below 135 mEq/L indicates hyponatremia that may hint to heart, kidney or liver problems, which notoriously have a negative impact on subjects with COVID-19 and pO2 values below 80 mmHg indicates that a person is not getting enough oxygen \cite{collins2015relating}.
The only binary decision of dubious interpretation from a medical perspective is the one based on glucose value at the third level of the tree. Probably, this is due to the overfitting problem that characterizes BDTs. However, this question and the following ones along its branches do not improve particularly the ability of the BDT to predict the risk of mortality. In fact, already before the debatable question on glucose the set was composed by 19 subjects who died and only 3 who survived.
In conclusion, overall the decisions taken by the BDT are medically grounded and the misclassified subjects amount only to the 5.5\% of the total population. Despite we recognised some overfitting effect, we believe that the 3-folded methodology presented in this paper can be successfully adopted to design an operative flowchart for clinical use.

\section{Conclusion}
Our work took place on the claim that causal learning, and in particular BSL methodologies, provide useful tools for delivering a robust and interpretable exploratory analysis of clinical data. We have put forward an integrated methodology, rooting on causal graph learning, that  provides practitioners with a pipeline of analytical steps amenable to
\begin{itemize}
    \item identify and measure the strength of causal association between features and outcomes in a fully-multivariate fashion;
    \item select causally relevant features and assess how they impact, separately, the outcome;
    \item provide a compact and interpretable procedural description of the outcome-determination process to support clinical decision making.
\end{itemize}

We have demonstrated the effectiveness of our approach on a COVID-19 case study, where we evaluated the causality relationship between clinical parameters and mortality in a cohort of COVID-19 patients. The analysis was carried out both splitting the clinical variables into classes as well as using them altogether into an integrated causal graph. Our clinical discussion provides compelling evidence of the robustness of the causal relationships identified by our causal learning approach. 
The results of the causal analysis were further used to build a highly explainable predictive model of mortality as a decision tree that could be easily implemented in a real clinical context. Our empirical validation against a permutation test setting, confirms the quality and relevance of the features identified by our method, also in a predictive setting. Thus, we suggest the adoption of this workflow to discover complex cause and effect relationships in clinical high dimensional dataset, also for other diseases. Within the context of this Coronavirus emergency, we believe that in future this multi-step approach may be used to better quantify the role of air pollution exposure \cite{travaglio2021links} with respect to other possible confounding variables and to investigate how the symptomatology, the care received and the clinical history of the patient impact on COVID long-term effects, which are still very unclear \cite{morlacco2020multifaceted,lopez2021more}. 
\FloatBarrier
\section*{Supporting information}
This section contains the results of the BSL analysis without incorporating any prior knowledge.
\FloatBarrier

\begin{figure}
  \begin{adjustwidth}{-2.25in}{0in}
    \centering
    \includegraphics[width=1.1\textwidth]{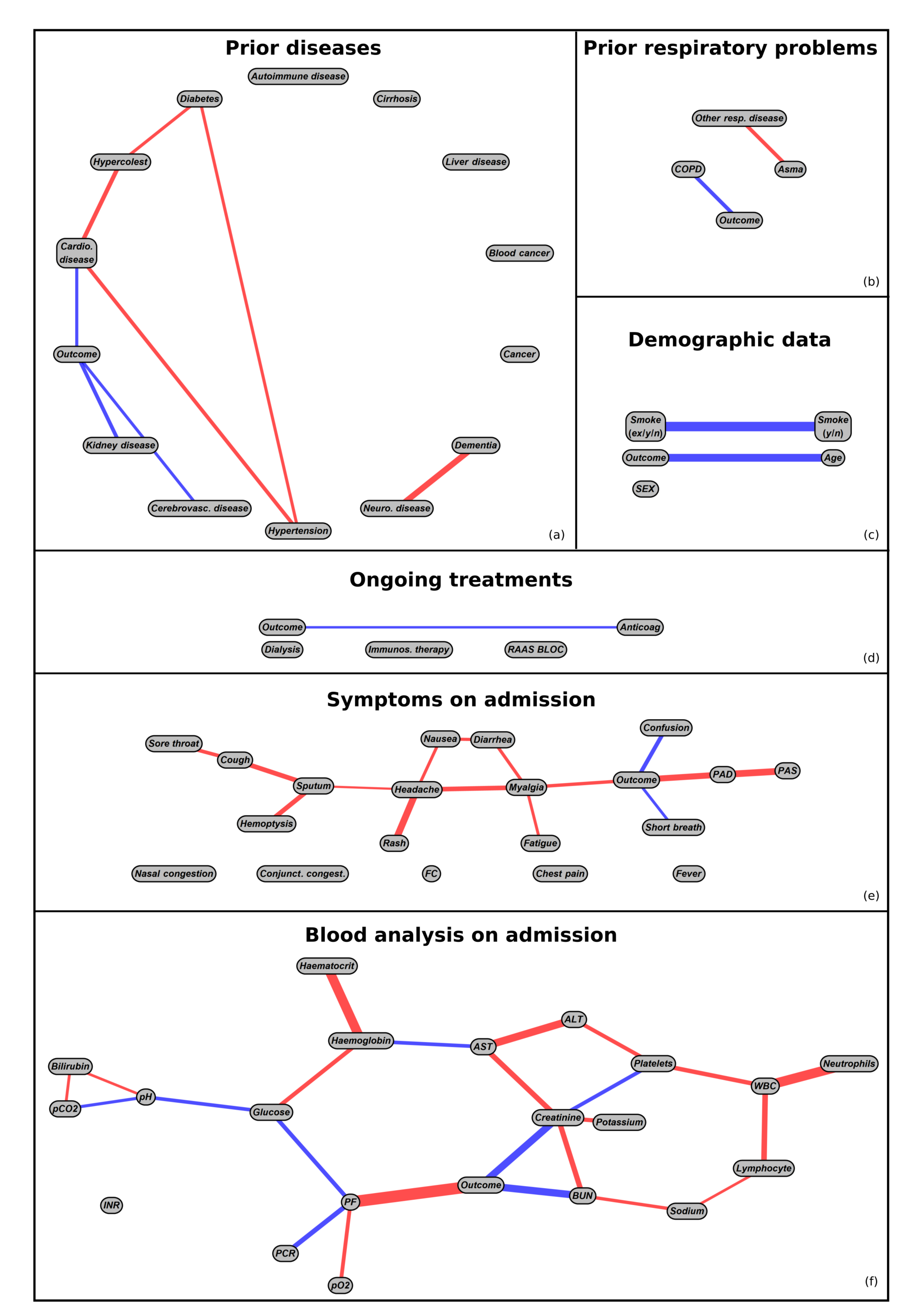}
    \caption{Image showing the BSL analysis applied to different categories of features. All the illustrated graphs are generated without taking the information provided by clinicians into account.}
    \label{fig_grafi_no_info}
  \end{adjustwidth}
\end{figure}

\begin{figure}
  \begin{adjustwidth}{-2.25in}{0in}
    \centering
    \includegraphics[width=1.1\textwidth]{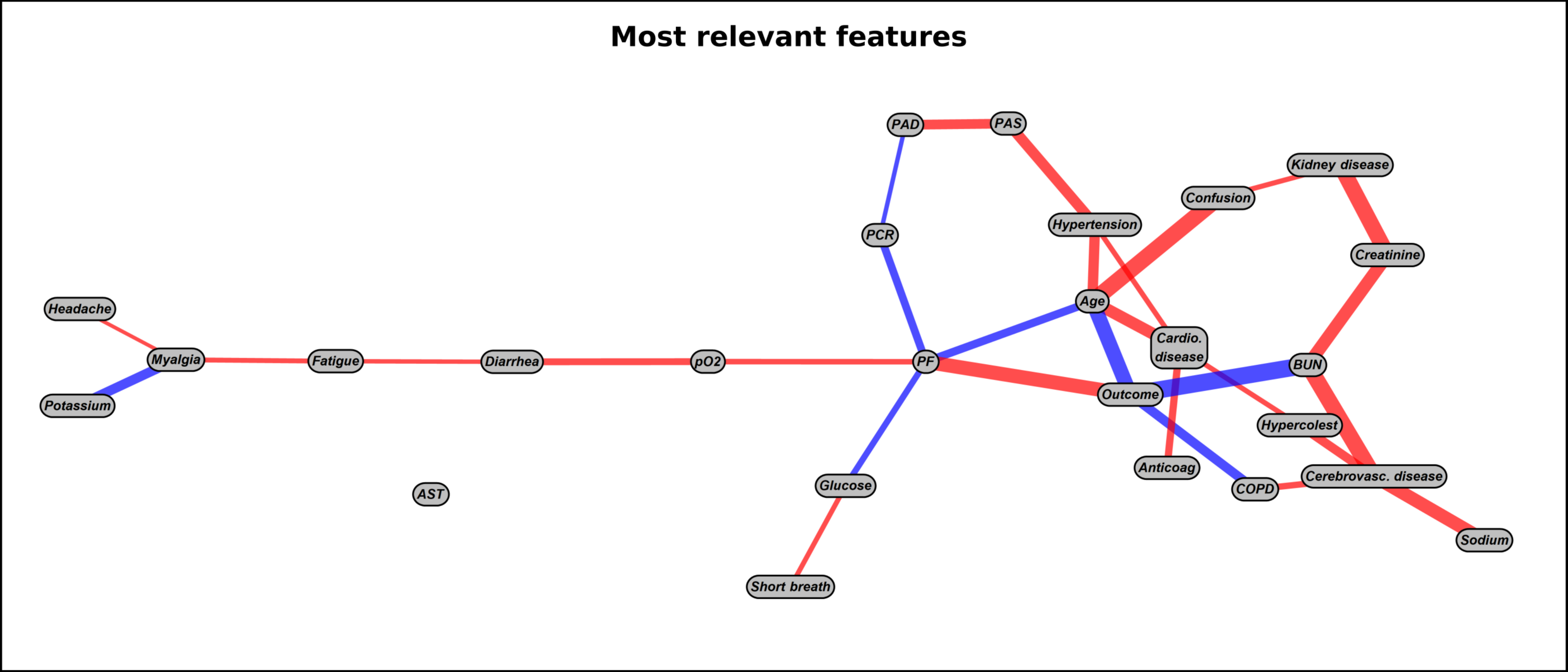}
    \caption{Graph generated with the most relevant features found from the graphs shown in Fig. \ref{fig_grafi_no_info}.}
    \label{fig_grafo_globale_no_info}
  \end{adjustwidth}
\end{figure}

\FloatBarrier


%
%
%
\bibliography{bibliography}

\begin{thebibliography}{10}

\bibitem{whoCovid}
{World Health Organisation}. WHO Coronavirus Disease (COVID-19) Dashboard.
  Accessed 8th March 2020; 2021.
\newblock Available from: \url{https://covid19.who.int/}.

\bibitem{kirkpatrick2020scarce}
Kirkpatrick JN, Hull SC, Fedson S, Mullen B, Goodlin SJ.
\newblock Scarce-resource allocation and patient triage during the COVID-19
  pandemic: JACC Review Topic of the Week.
\newblock Journal of the American College of Cardiology. 2020;76(1):85--92.

\bibitem{abd2020artificial}
Abd-Alrazaq A, Alajlani M, Alhuwail D, Schneider J, Al-Kuwari S, Shah Z, et~al.
\newblock Artificial Intelligence in the Fight Against COVID-19: Scoping
  Review.
\newblock Journal of medical Internet research. 2020;22(12):e20756.

\bibitem{wynants2020prediction}
Wynants L, Van~Calster B, Collins GS, Riley RD, Heinze G, Schuit E, et~al.
\newblock Prediction models for diagnosis and prognosis of covid-19: systematic
  review and critical appraisal.
\newblock bmj. 2020;369.

\bibitem{richens2020improving}
Richens JG, Lee CM, Johri S.
\newblock Improving the accuracy of medical diagnosis with causal machine
  learning.
\newblock Nature communications. 2020;11(1):1--9.

\bibitem{caruana2015intelligible}
Caruana R, Lou Y, Gehrke J, Koch P, Sturm M, Elhadad N.
\newblock Intelligible models for healthcare: Predicting pneumonia risk and
  hospital 30-day readmission.
\newblock In: Proceedings of the 21th ACM SIGKDD international conference on
  knowledge discovery and data mining; 2015. p. 1721--1730.

\bibitem{Spirtes2000}
Spirtes P, Glymour C, Scheines R.
\newblock Causation, Prediction, and Search.
\newblock 2nd ed. MIT press; 2000.

\bibitem{Aliferis2003371}
Aliferis CF, Statnikov AR, Tsamardinos I, Brown LE.
\newblock Causal Explorer: A Causal Probabilistic Network Learning Toolkit for
  Biomedical Discovery; 2003. p. 371--376.

\bibitem{miBN}
Bacciu D, Etchells TA, Lisboa PJ, Whittaker J.
\newblock Efficient Identification of Independence Networks Using Mutual
  Information.
\newblock Computational Statistics. 2013;28(2):621–646.

\bibitem{uusitalo2007advantages}
Uusitalo L.
\newblock Advantages and challenges of Bayesian networks in environmental
  modelling.
\newblock Ecological modelling. 2007;203(3-4):312--318.

\bibitem{heckerman2008tutorial}
Heckerman D.
\newblock A tutorial on learning with Bayesian networks.
\newblock Innovations in Bayesian networks. 2008; p. 33--82.

\bibitem{kontkanen1997comparing}
Kontkanen P, Myllym{\"a}ki P, Silander T, Tirri H, Grunwald P.
\newblock Comparing predictive inference methods for discrete domains.
\newblock In: In Proceedings of the sixth international workshop on artificial
  intelligence and statistics. Citeseer; 1997.

\bibitem{marcot2001using}
Marcot BG, Holthausen RS, Raphael MG, Rowland MM, Wisdom MJ.
\newblock Using Bayesian belief networks to evaluate fish and wildlife
  population viability under land management alternatives from an environmental
  impact statement.
\newblock Forest ecology and management. 2001;153(1-3):29--42.

\bibitem{walters2004fisheries}
Walters CJ, Martell SJ.
\newblock Fisheries ecology and management.
\newblock Princeton University Press; 2004.

\bibitem{imoto2003bayesian}
Imoto S, Kim S, Goto T, Aburatani S, Tashiro K, Kuhara S, et~al.
\newblock Bayesian network and nonparametric heteroscedastic regression for
  nonlinear modeling of genetic network.
\newblock Journal of bioinformatics and computational biology.
  2003;1(02):231--252.

\bibitem{jiang2011learning}
Jiang X, Neapolitan RE, Barmada MM, Visweswaran S.
\newblock Learning genetic epistasis using Bayesian network scoring criteria.
\newblock BMC bioinformatics. 2011;12(1):1--12.

\bibitem{van2014learning}
Van~der Heijden M, Velikova M, Lucas PJ.
\newblock Learning Bayesian networks for clinical time series analysis.
\newblock Journal of biomedical informatics. 2014;48:94--105.

\bibitem{onisko2016interpret}
Onisko A, Druzdzel MJ, Austin RM.
\newblock How to interpret the results of medical time series data analysis:
  classical statistical approaches versus dynamic Bayesian network modeling.
\newblock Journal of pathology informatics. 2016;7.

\bibitem{flores2011incorporating}
Flores MJ, Nicholson AE, Brunskill A, Korb KB, Mascaro S.
\newblock Incorporating expert knowledge when learning Bayesian network
  structure: a medical case study.
\newblock Artificial intelligence in medicine. 2011;53(3):181--204.

\bibitem{potere2020acute}
Potere N, Valeriani E, Candeloro M, Tana M, Porreca E, Abbate A, et~al.
\newblock Acute complications and mortality in hospitalized patients with
  coronavirus disease 2019: a systematic review and meta-analysis.
\newblock Critical care. 2020;24(1):1--12.

\bibitem{dana2020brazilian}
Dana S, Simas AB, Filardi BA, Rodriguez RN, da~Costa~Valiengo LL, Gallucci-Neto
  J.
\newblock Brazilian Modeling of COVID-19 (BRAM-COD): a Bayesian Monte Carlo
  approach for COVID-19 spread in a limited data set context.
\newblock MedRxiv. 2020;.

\bibitem{wibbens2020covid}
Wibbens PD, Koo WWY, McGahan AM.
\newblock Which COVID policies are most effective? A Bayesian analysis of
  COVID-19 by jurisdiction.
\newblock PloS one. 2020;15(12):e0244177.

\bibitem{mbuvha2020bayesian}
Mbuvha R, Marwala T.
\newblock Bayesian inference of COVID-19 spreading rates in South Africa.
\newblock PloS one. 2020;15(8):e0237126.

\bibitem{fenton2020privacy}
Fenton N, McLachlan S, Lucas P, Dube K, Hitman G, Osman M, et~al.
\newblock A privacy-preserving Bayesian network model for personalised COVID19
  risk assessment and contact tracing.
\newblock medRxiv. 2020;.

\bibitem{mclachlan2020fundamental}
McLachlan S, Lucas P, Dube K, McLachlan GS, Hitman GA, Osman M, et~al.
\newblock The fundamental limitations of COVID-19 contact tracing methods and
  how to resolve them with a Bayesian network approach.
\newblock London, UK https://doi org/1013140/RG. 2020;2(27042.66243).

\bibitem{neil2020bayesian}
Neil M, Fenton N, Osman M, McLachlan S.
\newblock Bayesian Network Analysis of Covid-19 data reveals higher Infection
  Prevalence Rates and lower Fatality Rates than widely reported.
\newblock Journal of Risk Research. 2020;23(7-8):866--879.

\bibitem{barbieri2020covid}
Barbieri G, Spinelli S, Filippi M, Foltran F, Giraldi M, Martino MC, et~al.
\newblock COVID-19 pandemic management at the Emergency Department: the
  changing scenario at the University Hospital of Pisa.
\newblock Emergency Care Journal. 2020;16(2).

\bibitem{beretta2018learning}
Beretta S, Castelli M, Gon{\c{c}}alves I, Henriques R, Ramazzotti D.
\newblock Learning the structure of Bayesian Networks: A quantitative
  assessment of the effect of different algorithmic schemes.
\newblock Complexity. 2018;2018.

\bibitem{buntine1996guide}
Buntine W.
\newblock A guide to the literature on learning probabilistic networks from
  data.
\newblock IEEE Transactions on knowledge and data engineering.
  1996;8(2):195--210.

\bibitem{daly2011learning}
Daly R, Shen Q, Aitken S.
\newblock Learning Bayesian networks: approaches and issues.
\newblock The knowledge engineering review. 2011;26(2):99.

\bibitem{scutari2019learns}
Scutari M, Graafland CE, Guti{\'e}rrez JM.
\newblock Who learns better Bayesian network structures: Accuracy and speed of
  structure learning algorithms.
\newblock International Journal of Approximate Reasoning. 2019;115:235--253.

\bibitem{spirtes2000causation}
Spirtes P, Glymour CN, Scheines R, Heckerman D.
\newblock Causation, prediction, and search.
\newblock MIT press; 2000.

\bibitem{spirtes1991algorithm}
Spirtes P, Glymour C.
\newblock An algorithm for fast recovery of sparse causal graphs.
\newblock Social science computer review. 1991;9(1):62--72.

\bibitem{finney1948fisher}
Finney DJ.
\newblock The Fisher-Yates test of significance in 2$\times$ 2 contingency
  tables.
\newblock Biometrika. 1948;35(1/2):145--156.

\bibitem{kornbrot2005point}
Kornbrot D.
\newblock Point biserial correlation.
\newblock Encyclopedia of statistics in behavioral science. 2005;.

\bibitem{bramer2007avoiding}
Bramer M.
\newblock Avoiding overfitting of decision trees.
\newblock Principles of data mining. 2007; p. 119--134.

\bibitem{sanchez2020underlying}
Sanchez-Ramirez DC, Mackey D.
\newblock Underlying respiratory diseases, specifically COPD, and smoking are
  associated with severe COVID-19 outcomes: A systematic review and
  meta-analysis.
\newblock Respiratory medicine. 2020; p. 106096.

\bibitem{gacche2021predictors}
Gacche R, Gacche R, Chen J, Li H, Li G.
\newblock Predictors of morbidity and mortality in COVID-19.
\newblock European review for medical and pharmacological sciences.
  2021;25(3):1684--1707.

\bibitem{xu2020association}
Xu J, Xiao W, Liang X, Zhang P, Shi L, Wang Y, et~al.
\newblock The association of cerebrovascular disease with adverse outcomes in
  COVID-19 patients: a meta-analysis based on adjusted effect estimates.
\newblock Journal of Stroke and Cerebrovascular Diseases. 2020;29(11):105283.

\bibitem{fathi2021prognostic}
Fathi M, Vakili K, Sayehmiri F, Mohamadkhani A, Hajiesmaeili M, Rezaei-Tavirani
  M, et~al.
\newblock The prognostic value of comorbidity for the severity of COVID-19: A
  systematic review and meta-analysis study.
\newblock PloS one. 2021;16(2):e0246190.

\bibitem{sharma2020liver}
Sharma A, Jaiswal P, Kerakhan Y, Saravanan L, Murtaza Z, Zergham A, et~al.
\newblock Liver disease and outcomes among COVID-19 hospitalized patients-a
  systematic review and meta-analysis.
\newblock Annals of hepatology. 2020;.

\bibitem{coppelli2020hyperglycemia}
Coppelli A, Giannarelli R, Aragona M, Penno G, Falcone M, Tiseo G, et~al.
\newblock Hyperglycemia at hospital admission is associated with severity of
  the prognosis in patients hospitalized for COVID-19: the Pisa COVID-19 Study.
\newblock Diabetes Care. 2020;43(10):2345--2348.

\bibitem{july2021prevalence}
July J, Pranata R.
\newblock Prevalence of dementia and its impact on mortality in patients with
  coronavirus disease 2019: A systematic review and meta-analysis.
\newblock Geriatrics \& Gerontology International. 2021;21(2):172--177.

\bibitem{yadaw2020clinical}
Yadaw AS, Li Yc, Bose S, Iyengar R, Bunyavanich S, Pandey G.
\newblock Clinical features of COVID-19 mortality: development and validation
  of a clinical prediction model.
\newblock The Lancet Digital Health. 2020;2(10):e516--e525.

\bibitem{peckham2020male}
Peckham H, de~Gruijter NM, Raine C, Radziszewska A, Ciurtin C, Wedderburn LR,
  et~al.
\newblock Male sex identified by global COVID-19 meta-analysis as a risk factor
  for death and ITU admission.
\newblock Nature communications. 2020;11(1):1--10.

\bibitem{zhang2021association}
Zhang H, Ma S, Han T, Qu G, Cheng C, Uy JP, et~al.
\newblock Association of smoking history with severe and critical outcome in
  COVID-19 patients: A systemic review and meta-analysis.
\newblock European journal of integrative medicine. 2021; p. 101313.

\bibitem{austad2006women}
Austad SN.
\newblock Why women live longer than men: sex differences in longevity.
\newblock Gender medicine. 2006;3(2):79--92.

\bibitem{zheng2020risk}
Zheng Z, Peng F, Xu B, Zhao J, Liu H, Peng J, et~al.
\newblock Risk factors of critical \& mortal COVID-19 cases: A systematic
  literature review and meta-analysis.
\newblock Journal of Infection. 2020;.

\bibitem{qiu2020clinical}
Qiu P, Zhou Y, Wang F, Wang H, Zhang M, Pan X, et~al.
\newblock Clinical characteristics, laboratory outcome characteristics,
  comorbidities, and complications of related COVID-19 deceased: a systematic
  review and meta-analysis.
\newblock Aging clinical and experimental research. 2020; p. 1--10.

\bibitem{cheng2020kidney}
Cheng Y, Luo R, Wang K, Zhang M, Wang Z, Dong L, et~al.
\newblock Kidney impairment is associated with in-hospital death of COVID-19
  patients.
\newblock MedRxiv. 2020;.

\bibitem{ware2000acute}
Ware LB, Matthay MA.
\newblock The acute respiratory distress syndrome.
\newblock New England Journal of Medicine. 2000;342(18):1334--1349.

\bibitem{collins2015relating}
Collins JA, Rudenski A, Gibson J, Howard L, O’Driscoll R.
\newblock Relating oxygen partial pressure, saturation and content: the
  haemoglobin--oxygen dissociation curve.
\newblock Breathe. 2015;11(3):194--201.

\bibitem{travaglio2021links}
Travaglio M, Yu Y, Popovic R, Selley L, Leal NS, Martins LM.
\newblock Links between air pollution and COVID-19 in England.
\newblock Environmental Pollution. 2021;268:115859.

\bibitem{morlacco2020multifaceted}
Morlacco A, Motterle G, Zattoni F.
\newblock The multifaceted long-term effects of the COVID-19 pandemic on
  urology.
\newblock Nature Reviews Urology. 2020;17(7):365--367.

\bibitem{lopez2021more}
Lopez-Leon S, Wegman-Ostrosky T, Perelman C, Sepulveda R, Rebolledo PA, Cuapio
  A, et~al.
\newblock More than 50 Long-term effects of COVID-19: a systematic review and
  meta-analysis.
\newblock Available at SSRN 3769978. 2021;.

\end{thebibliography}
\end{document}